\documentclass{iaf} 

\usepackage{times}

\usepackage{latexsym} 

\usepackage{graphics}
\usepackage{epsf}
\usepackage{float}

\newcommand{\poub}[1]{}

\newcommand{\isaobj}{\ensuremath{\stackrel{\mbox{\tiny is-a}}{\longrightarrow}}~}
\newcommand{\isaont}{\ensuremath{\stackrel{\mbox{\tiny ded ont}}{\longrightarrow}}~} 
\newcommand{\isaontset}{\ensuremath{\rightarrow^{\mbox{\tiny ded ont}}_{\mbox{\tiny set}}}~} 
\newcommand{\cause}{~\ensuremath{causes}~}

\newcommand{\hfilld}{\hfill\makebox[0em]{}}

\newcommand{\cexpl}{\textit{~can explain~}}

\newcommand{\cbexpl}{\mbox{~\it can be explained by~}}
\newcommand{\prov}[1]{\textit{~provided~ {\it #1} is possible}}

\newtheorem{defi}{Definition}

\annee{2013}
\titre{Arguments using ontological and causal knowledge}
\auteurs{Philippe Besnard$^1$$^4$ \and 
Marie-Odile Cordier$^2$$^4$  \and 
Yves Moinard$^3$$^4$} 

\institutions{
$^1${ CNRS / IRIT, Universit\'{e} de Toulouse, France}\\
$^2${ Universit\'{e} de Rennes I}\\
$^3${ INRIA Bretagne Atlantique}\\
$^4${ IRISA, Campus  de Beaulieu, 35042 Rennes Cedex, France}}
 \mels{besnard@irit.fr, cordier@irisa.fr, yves.moinard@inria.fr}

\begin{document}
\creationEntete

\begin{resume}
Nous d\'ecrivons l'utilisation d'un syst\`eme logique de raisonnement \`a partir de donn\'ees causales
et ontologiques dans un cadre argumentatif. 
Les donn\'ees consistent en liens {\em causaux} ({\em $\{A_1, \cdots ,A_n\}$ cause B}) et  {\em ontologiques} ($o_1$ {\em est\_un} $o_2$). Le syst\`eme en d\'eduit des {\em liens explicatifs} possibles ($\{A_1, \cdots ,A_n\}$ {\em explique $\{B_1, \cdots ,B_m\}$}). Ces liens explicatifs servent ensuite de base \`a un syst\`eme argumentatif qui fournit des {\em explications} possibles.
Un exemple inspir\'e de la temp\^ete Xynthia, laquelle a provoqu\'e un trop grand nombre de victimes  par rapport aux conditions purement m\'et\'eorologiques, illustre une utilisation de notre syst\`eme.
\end{resume}

\begin{abstract}
We investigate an approach to reasoning about causes through argumentation. 
We consider a causal model for a physical 
system, and look for arguments about  facts. 
Some arguments are meant to provide explanations of facts whereas some challenge these explanations and so on. 
At the root of argumentation here, are causal links ({\em $\{A_1, \cdots ,A_n\}$ causes B})
and ontological links ($o_1$ {\em is\_a} $o_2$). 
We present a system that provides a candidate explanation 
($\{A_1, \cdots ,A_n\}$ {\em explains $\{B_1, \cdots ,B_m\}$}) by resorting to an underlying causal link substantiated with appropriate ontological links. 
Argumentation is then at work from these various explaining links.
A case study is developed: a severe storm  Xynthia that devastated part of France in 2010, with an unaccountably high number of casualties. 

\end{abstract}

\section{Introduction and Motivation} 
 
Looking for explanations is a frequent operation, in various domains, from judiciary to mechanical fields. We consider the case where we have some precise (not necessarily exhaustive) description of some mechanism, or situation, and we are looking for explanations of some facts. The description contains logical formulas, plus some {\em causal} and {\em ontological} formulas (or links). Indeed, it is known that, while there are similarity between {\em causation} and {\em implication}, causation cannot be rendered by a simple logical implication.  
Also, confusing causation and {\em co-occurrence} could lead to undesirable relationships.  
This is why we use here a {\em causal formalism} where some {\em causal links} and {\em ontological links} are added to classical logical formulas. Then the causal formalism will produce various {\em explanation links} \cite{BesCorMoi08a}. However, if the situation described is complex enough, this will result in a great number of possible explanations, and some {\em argumentation} is involved in order to get some reasons to choose between all these candidate explanations.  

In this text, we will consider as an example a severe storm, called Xynthia, which made 26 deaths in a single group of houses in La Faute sur Mer, a village  
in Vend\'{e}e during a night in February 2010.  
This was a severe storm, with strong winds, low pressure, but it had been forecast. Since the casualties were excessive with respect to the strength of the meteorological phenomenon, various investigations have been ordered. This showed that various factors combined their effects. The weather had its role, however, other factors had been involved: recent houses and a fire station had been constructed in an area known as being susceptible of getting submerged. 
Also, the state authorities did not realize that asking people to stay at home was inappropriate in case of flooding given the traditionally low Vend\'ee houses. 

In this paper, we define in section \ref{secenrich} an enriched causal model, built from a causal model and an ontological model. We then show in section \ref{secexpl} how explanations can be derived from this enriched causal model. We explain in section \ref{secarg} the use of argumentation in order to deal with the great number of possible explanations obtained and conclude in section \ref{secconcl}. The Xynthia example, introduced in section \ref{secXyn}, is used as illustration throughout the article.

\section{Enriched causal model = Causal model + ontological model}\label{secenrich} 
The model that is used to build tentative explanations and support argumentation, called the enriched causal model, is built from a causal model relating literals in causal links, and from an ontological model where classes of objects are related through specialization/generalization links.

\subsection{The causal model}\label{sseccaus} 
By a causal model \cite{Mel95}, we mean a representation of a body of causal relationships to be used to generate arguments that display explanations for a given set of facts. 

The basic case is that of a causal link ``$\alpha$ causes $\beta$'' where $\alpha$ and $\beta$ are literals. 
In this basic case, $\alpha$ stands for the singleton $\{\alpha\}$ as the general case of 
a {\em causal link} is  the form \[\{\alpha_1, \alpha_2,\cdots,\alpha_n\} \cause \beta\]
 where $\{\alpha_1, \alpha_2,\cdots \alpha_n\}$ is a set (interpreted conjunctively) of literals. 
 
Part of the causal model for our Xynthia example is given in Fig.~\ref{gddessinC} 
(each plain black arrow represents a causal link).

\subsection{The ontological model}\label{ssecontol}

The literals $P(o_1,o_2,\cdots,o_k)$ occurring in the causal model use some predicates $P$ applied to classes of objects $o_i$. The ontological model consists of specialization/generalization links between classes of objects \[o_1 \isaobj o_2,\]
 where $\isaobj$ denotes the usual specialization link between classes. E.g., we have {\tt Hurri} \isaobj {\tt SWind}, {\tt House1FPA} \isaobj {\tt HouseFPA} and {\tt HouseFPA} \isaobj {\tt BFPA}: a ``hurricane'' ({\tt Hurri}) is a specialization of a ``strong wing'' ({\tt SWind}), and 
the class of ``low houses with one level only in the flood-prone area'' ({\tt House1FPA}) is a specialization of the class of ``houses in the flood-prone area'' ({\tt HouseFPA}), which itself is a specialization of the class of
``buildings in this area'' ({\tt BFPA}). A part of the ontological model for our Xynthia example is given in Fig.~\ref{gddessinO} 
(each white-headed arrow labelled with {\em is-a}  represents an \isaobj link).

\subsection{The enriched causal model}\label{ssecenrich} 
 
The causal model is extended by resorting to the ontological model, and the result is called the enriched causal model. 
The enrichment lies in the so-called {\em ontological deduction links} (denoted \isaont) between literals. 
Such a link \[\alpha \isaont \beta\] simply means that $\beta$ can be deduced from $\alpha$ due to 
specialization/generalization links, the \isaobj links in the ontological model, that relate the classes of objects mentioned in $\alpha$ and $\beta$. Note that the relation \isaont is transitive and reflexive. 
 
Here is an easy illustration. 
A sedan is a kind of car, which is represented by $sedan \isaobj car$ in the ontological model. 
Then, $Iown(sedan) \isaont Iown(car)$ is an ontological deduction link in the enriched model.
 
Technically, the \isaont links among literals are generated by means of a 
single principle as follows. 
The predicates used in the causal model are annotated so that each of their parameters is sorted either as {\em universal} or as {\em existential}.
 
A universal parameter of a predicate ``inherits'' by specialization, meaning that if the predicate is true on this parameter then the predicate is also true for specializations of this parameter. 
The existential parameters of a predicate ``inherits'' by generalization, meaning that if a the predicate is true on this parameter, then the predicate is also true for generalizations of this parameter (cf above example of our owner of a sedan). 

As another example, consider the unary predicate {\tt Flooded}, where {\tt Flooded$\!$($\!$o$\!$)} means that class {\tt o} (an area or a group of buildings) is submerged. 
Its unique parameter is taken to be ``universal'' so that if the literal {\tt Flooded$\!$($\!$o$\!$)} is true, then the literal {\tt Flooded$\!$($\!$s$\!$)} is also true whenever $s$ is a specialization of $o$. 
The causal model is enriched by adding {\tt Flooded$\!$($\!$o$\!$)} \isaont {\tt Flooded$\!$($\!$s$\!$)} for each class $s$ satisfying $s \isaobj o$. 

Let us now consider the unary predicate {\tt Occurs} so that {\tt Occurs(Hurri)} intuitively means: 
some hurricane occurs. Exactly as above ``I own'' predicate, this predicate is existential on its unique parameter. 
By means of the \isaobj link {\tt Hurri} \isaobj {\tt SWind}, we obtain the following \isaont link:\ 
{\tt Occurs(Hurri)} \isaont {\tt Occurs(SWind)}. 

Let us provide the formal definition of \isaont links in the case of unary predicates. Then, the general case is a natural, but intricate and thus ommitted here, generalization
of the unary case.

\begin{defi}\label{defdedont}{\rm 
Let us suppose that $Prop1_\exists$ and  $Prop2_\forall$ are two unary predicates, of the ``existential kind'' for the first one, and of the ``universal kind'' for the second one.
If  in the ontology is the link
$class_1\isaobj class_2$, 
then the following two links are added to the enriched model:\\
$ Prop1_\exists(class_1) \isaont  Prop1_\exists(class_2)$ and\\
$ Prop2_\forall(class_2) \isaont  Prop2_\forall(class_1)$.
}\end{defi}

In our formalism, causal and ontological links \isaont entail classical implication:
\begin{equation}\label{EC}
\begin{array}{rcl}\{\alpha_1,\cdots \alpha_n\}\cause\alpha & \mbox{entails} &(\bigwedge_{i=1}^n\! \alpha_i) \rightarrow \alpha.\\
\alpha \isaont \beta &\mbox{entails}& \alpha \rightarrow \beta. 
\end{array}\end{equation}%
Ordinary logical formulas are also allowed (for example for describing exclusions),
which are added to the formulas coming from (\ref{EC}).

When resorting to explanations (see section \ref{secexpl} below),  
these \isaont links are  extended to sets of literals (links denoted by \isaontset) as follows: 
\begin{defi}\label{defdedontset}{\rm
 Let $\Phi$ and $\Psi$ be two sets of literals, we define
 $\Phi \isaontset  \Psi$, if for each $\psi\in\Psi$, there exists 
$\varphi\in\Phi$ such that $\varphi\isaont\psi$ (remind that $\psi\isaont\psi$). 
}\end{defi}
Notice that if a sedan is a kind of car ($sedan \isaobj car$), and $Own$ a predicate existential on its unique argument, then we get\hfill
$\{Iown(sedan),IamHappy\}$ \isaontset\hfilld\\\makebox[6.5em]{} $\{Iown(car),IamHappy\}$, \hfill and\\ 
\makebox[2em]{}$\{Iown(sedan),IamHappy\}$ \isaontset $\{Iown(car)\}.$

\section{The Xynthia example} \label{secXyn} 
 
From various enquiries, including one from the French parliament%
\footnote{\tt http://www.assemblee-nationale.fr/13/rap-info/\\ i2697.asp} 
and one from the Cours des Comptes%
\footnote{\tt www.ccomptes.fr/Publications/Publications/\\Les-enseignements-des-inondations-de-2010-sur\\-le-littoral-atlantique-Xynthia-et-dans-le-Var} and many articles on the subject, we have many information about the phenomenon and its dramatic consequences. 
We have extracted a small part from all the information 
as an illustration of our approach. 
 
The classes we consider in the ontological model and/or in the causal model are the following ones~:  {\tt Hurri}, {\tt SWind}, {\tt BFPA}, {\tt House1FPA}, {\tt HouseFPA}, and {\tt BFPA}
have already been introduced in \S  \ref{ssecontol}, together with a few \isaobj links.
 Among the buildings in the flood-prone area, there is also a fire station {\tt FireSt} (remind that we have also a group of houses {\tt HouseFPA}, including a group of typical Vend\'ee low houses with one level only
 {\tt House1FPA}).
We consider also three kinds of natural disasters ({\tt NatDis}): {\tt Hurri}, together with ``tsunami'' ({\tt Tsun}) and ``flooding'' ({\tt Flooding}).
As far as meteorological phenomena are concerned, we restrict ourselves to ``Very low pressure'' ({\tt VLPress}), together with already seen {\tt Hurri} and {\tt SWind},
and finally we add `high spring tide'' ({\tt HSTide}) to our list of classes.

Two kinds of alerts ({\tt Alert}) may be given by the authorities, ``Alert-Evacuate'' ({\tt AlertEvac}) and ``Alert-StayAtHome'' ({\tt AlertStay}). 
We consider also an anemoter ({\tt Anemo}) able to measure the wind strength and a fact asserting that ``people stay at home'' {(\tt PeopleStay}). 
 
The following predicates are introduced: ({\tt Flooded}) and  ({\tt Victims\_I}) applied to a group of building respectively meaning that ``flooding'' 
occurs over this group, and that there were ``victims'' in this group ($I\in\{1,2,3\}$ is a degree of gravity, e.g. {\tt Victims\_1}, {\tt Victims\_2} and {\tt Victims\_3} respectively mean, in \% of the population of the group: 
``a small number'', ``a significant number'' and ``a large number'' of victims). \label{VictimsI}  
{\tt OK} means that its unique parameter is in a normal state. 
{\tt Occurs} means that some fact has occurred (a strong wind, a disaster, \ldots), {\tt Expected} means that some fact is expected to occur.\\ 

All these  predicates are ``universal'' on their unique parameter, except for the predicates {\tt Occurs} and {\tt Expected} which are ``existential''.\\ 

Note that negation of a literal is expressed by {\tt Neg} as in {\tt Neg OK(Anemo)} meaning that the anemometer is not in its normal behaviour state (the formula $\neg {\tt(OK(Anemo)} \wedge {\tt(NEG\;OK(anemo))}$ is thus added). \\

The classes and the ontological model are given in Figure \ref{gddessinO} with the  \isaobj links represented as white-headed arrows labelled with {\tt (is-a)}. 

\begin{samepage}\begin{figure}[!h]
\centerline{\includegraphics{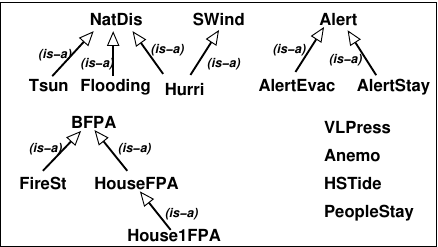}}
\vspace*{-1ex} \caption{Ontological model for Xynthia}\label{gddessinO} 
\end{figure}\end{samepage}

\begin{figure}[!h]
\centerline{\includegraphics{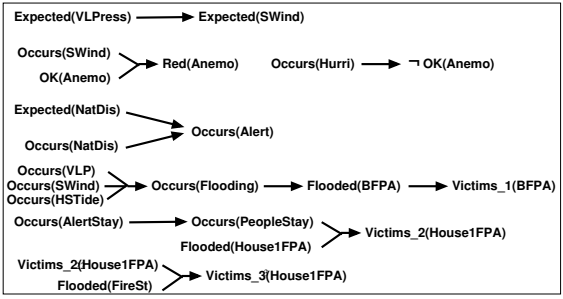}} 
\caption{A part of the causal model for Xynthia}\label{gddessinC} 
\end{figure}

Part of the causal model is given in Figure \ref{gddessinC} (remember, black-headed arrows represent causal links). It represents  causal relations between [sets of] 
literals. 
It expresses that an alert occurs when a natural disaster is expected, or when a natural disaster occurs. Also, people stay at home if alerted to stay at home, and  having one level home  flooded results  in many victims, and even more victims if the fire station itself is flooded,...\\
 
From the ontological model and the causal model, the enriched causal model can be build, adding \isaont 
links between literals when possible. 

For instance, for our ``existential'' predicates, from {\tt Hurri} \isaobj {\tt SWind},  the links\\ {\tt Occurs(Hurri)} \isaont  {\tt Occurs(SWind)}, and\\
  {\tt Expected(Hurri)} \isaont { \tt Expected(SWind)} are added. 

In the same way, for a universal predicate, from {\tt House1FPA} \isaobj {\tt BFPA}, 
is added 
the link \mbox{{\tt Flooded(BFPA)} \isaont {\tt Flooded(House1FPA)}} (cf Fig. \ref{figvictims1}).\\

\begin{figure}[!h]
\centerline{\includegraphics{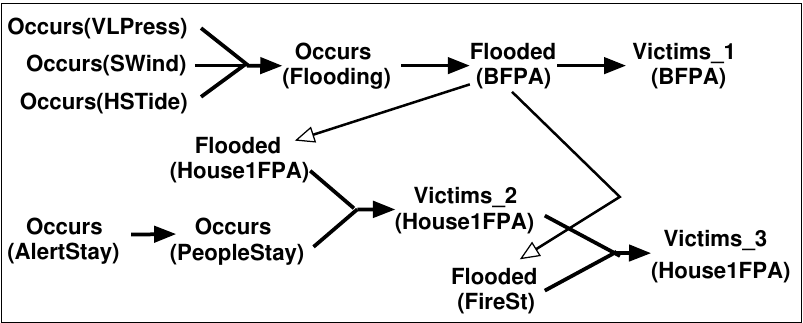}
}
\caption{Numerous victims in low and flood-prone houses\vspace*{-5ex}}
\label{figvictims0}
\end{figure}

Figure \ref{figvictims0} represents a part of the enriched causal model, where the (unlabelled) white-headed arrows represent the \isaont links and each black-headed arrow represents a causal link (from a literal or, in case of forked entry,  form a set of literals). 

\section{Explanations} \label{secexpl}
\subsection{Explaining a literal from a [set of] literal[s]}\label{ssecexpl} 
Causal and ontological links allow us to infer {\em explanation links}.
We want to exhibit candidate reasons that can explain 
a fact by means of at least one causal link.
We disregard explanations that involve only links of the implicational kind.
Here is how causal and ontological links are used in our formalism:

Let $\Phi$ denote a set of literals 
and $\beta$ be a literal. 
\begin{equation}\label{eqbasic}
\begin{array}{l}\textrm{The basic case is that }\hspace{1em} \Phi \cause \beta \\
 \textrm{yields that}\hspace{5em} 
\beta \textrm{  can be explained by } \Phi.\end{array}%
\end{equation}

The general initial case involves two literals 
$\beta=Prop(cl_2)$ and $\delta=Prop(cl_1)$
built on the same predicate $Prop$
(eventual other parameters equal in these two literals):\vspace{-1ex}
\begin{equation}\label{eqC1}
\makebox[-1em]{}\left\{\makebox[-.5em]{}\begin{array}{c}\Phi \cause \beta \\ \delta \isaont \beta \end{array}\makebox[-.5em]{}\right\}
\mbox{yields that}
\begin{array}{l} \delta \textit{ can be explained by } \Phi,\\
 \textit{ provided } \delta \textit{ is possible.}
\end{array}\vspace{-1ex}
\end{equation}

\noindent$\{\delta\}$ is the {\em set of justifications} for this explanation of $\delta$ by $\Phi$.\\

 We require also that the origin of the explanation is possible, thus the {\em full set of justifications} is here the set $\Phi \cup \{\delta\}$, while in case of (\ref{eqbasic}), this full set is $\Phi$, even if no set is given explicitly.
Indeed, it is always understood that the starting point of the explanation link must be possible, thus we sometimes omit it.
Remind that from (\ref{EC}) we get $(\bigwedge_{\varphi\in\Phi}\varphi) \rightarrow\beta$ and $\delta \rightarrow \beta$, thus 
adding $\beta$ to the justification set is useless.\\

An explaining link\\
 $\Phi \cexpl \alpha \prov{$\Psi$}$ \\
 where $\Psi$ is a set of literals means that\\
$\alpha \cbexpl \Phi$ provided the set $\Phi \cup \Psi$ is possible:\\ if adding $\Phi \cup \Psi$ to the available data leads to inconsistency, then the explanation link cannot be used
to explain $\delta$ by $\Phi$.
\\

In the figures, white-headed arrows represent ``ontological links'': \isaont links for bare arrows, and \isaobj links for arrows with {\tt (is-a)} mentioned, while 
dotted arrows represent explanation links (to be read {\em can explain}) produced by these rules, these arrows being sometimes labelled with the corresponding justification set. 

\begin{figure}[!h]
\centerline{\includegraphics{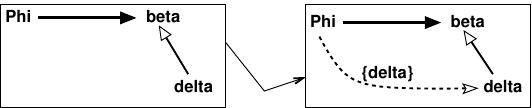}}
\caption{The explanation link from (\ref{eqC1}) when $\Phi=\{\varphi\}$}\vspace{-1ex}
\label{figexPhilip1}
\end{figure}

We want also that our explanation links can follow \isaont links as follows, in the case where 
$\Phi=\{\varphi\}$ is a singleton (thus we note sometimes
$\varphi$ instead of $\{\varphi\}$):%
\begin{equation}\label{eqC2}
\makebox[-1em]{}\begin{array}{c}
\left\{\begin{array}{c}\varphi \cexpl \delta \prov{$\Psi$}\\ \varphi_0\isaont \varphi \hspace{2em}\delta \isaont\delta_1\end{array}\right\}\\[2ex]
\mbox{yields that } 
 \varphi_0 \cexpl \delta_1 \prov{$\Psi$}
\end{array}
\end{equation}

\noindent$\{\varphi_0\} \cup \Psi$ is the {\em full} justification set for this explanation of $\delta_1$ by $\varphi_0$. Again we get $\varphi_0 \rightarrow \varphi$ 
by (\ref{EC}), so we need not to mention $\varphi$ here.  

\begin{figure}[!h]
\centerline{\includegraphics{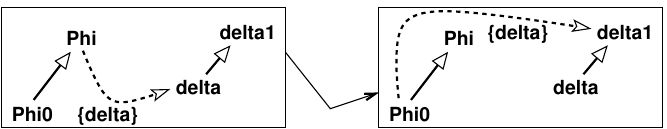}
}
\caption{Explanation links follow \isaont links [cf (\ref{eqC2})]}
\label{figexpldedont}
\end{figure}

Generalizing the case of (\ref{eqC2}) when $\Phi$ is a set of literals needs the use of \isaontset:%
\begin{equation}\label{eqC2set}
\makebox[-1em]{}\begin{array}{c}
\left\{\begin{array}{c}\Phi \cexpl \delta \prov{$\Psi$}\\ \Phi_0\isaontset \Phi \hspace{2em}\delta \isaont\delta_1\end{array}\right\}\\[2ex]

\mbox{yields that } 
 \Phi_0 \cexpl \delta_1 \prov{$\Psi$}
\end{array}
\end{equation}

Now, we also want that our explanation links are transitive, and this necessitates to be able to explain not only literals, but sets of literals.
Thus we introduce explanation links among sets of literals, which extend our explanation links from sets of literals towards literals (since it is an extension, we can keep the same name {\em explanation link}):

\subsection{Explaining a set of literals from a set of literals}\label{ssecexplset} 

\begin{defi}\label{defexplset0}
{\rm 
Let $n$ be some natural integer, and, for $i\in \{1,2,\cdots, n\}$ $\Phi, \Psi_i$ be sets of literals and $\delta_i$ be literals.
If, for each $i\in \{1,2,\cdots, n\}$, we have
$\Phi \cexpl \delta_i \prov{$\Psi_i$},$\vspace{1ex}\\
then we define the following {\em [set] explanation link}\vspace{1ex}\\
$\begin{array}{r} \Phi \cexpl \{\delta_i / i\in \{1,2,\cdots, n\}\}\\
  \prov{$\bigcup_{i=1}^n \Psi_i$}.\end{array}$\vspace{1ex}\\

Again, such an explanation link applies only when its (full) justification set, here $\Phi \cup\bigcup_{i=1}^n \Psi_i$,
is possible (not contradicted by the data).
}\end{defi}

Notice that we do not want to explain $\Phi$ by $\Phi$ itself, and we extend this restriction to \isaontset links:

We do not want to explain $\Phi$ by $\Phi_0$ is all we know is
$\Phi_0 \isaontset \Phi$.  Indeed, this seems to be cheating about what an explanation is (we want some causal information to play a role).

However, in the line of (\ref{eqC2set}), we want that explanations follow \isaontset links, thus we introduce the following definition:

\begin{defi}\label{defexplsetisa}{\rm
If we have\vspace{1ex}\\
$\begin{array}{c}
\noindent\left\{\begin{array}{c}\Phi \cexpl \Delta \prov{$\Psi$}\\ \Phi_0\isaontset \Phi \hspace{2em}\Delta \isaontset\Delta_1\end{array}\right\}\vspace{1ex}\\
\mbox{then we have } \vspace{1ex}\\
 \Phi_0 \cexpl \Delta_1 \prov{$\Psi$}\end{array}$\vspace{1ex}\\
Again, the full justification set of the resulting explaining link is $\Phi_0 \cup\Psi$.
}\end{defi}

Our last definition of [set] explanation links concerns transitivity of explanations.
This is a ``weak'' transitivity since the justifications are gathered, however, we will call this property ``transitivity''.
We need to be able 
to omit in the resulting link the part of the intermediate set which is already explained in the first  explanation giving rise to a transitive link:

\begin{defi}\label{defexplsettr}
{\rm 
If \\
$\left\{\hspace{-1ex}\begin{array}{rlc}
\Phi \cexpl \Delta_1 \cup \Delta_2 \prov{$\Psi_1$} &\textrm{and}\\
\Gamma \cup \Delta_2 \cexpl \Theta \prov{$\Psi_2$},\end{array}\hspace{-1ex}\right\}$\vspace{1ex}\\
then \hspace{.5em} 
$\Phi \cup \Gamma  \cexpl \Delta_1 \cup \Delta_2 \cup \Theta\\
\makebox[3em]{} \prov{$\Psi_1 \cup \Psi_2$}.$\\

Again, the full justification set is $\Phi \cup \Gamma \cup \Psi_1\cup \Psi_2$.
}
\end{defi}

\subsection{About explanation links and arguments}

As a small example, let us suppose that in the causal model is the link

$\mbox{reason~} \cause Prop_\exists(class_2)$

and that in the ontology are the links:

$\begin{array}{c}class_1\isaobj class_2 \\ class_1\isaobj class_3\end{array}$\\

Resulting are the next two links in the enriched model:

$\begin{array}{c}
Prop_\exists(class_1) \isaont  Prop_\exists(class_2)\\
Prop_\exists(class_1) \isaont  Prop_\exists(class_3)\end{array}$\\

By pattern matching over the diagrams in Fig.~\ref{figexPhilip1} and \ref{figexpldedont} [cf patterns (\ref{eqC1}) and (\ref{eqC2set}], we get

\begin{center}
reason~can explain $Prop_\exists(class_3)$\hfill through\hfill $Prop_\exists(class_1)$\hfill(``justification'').
\end{center}
Importantly, it is assumed that the causal link is expressed on the appropriate level: 
in other words, should there be some doubts about the kind of objects (here $class_2$) that enjoy $Prop_\exists$ due to \emph{reason}, the causal link would be about another class.

The proviso accompanying the explanation takes place at the level of justifications: 
the candidate explanation is worth inasmuch as $Prop_\exists(class_1)$ is not contradicted. 
In particular it must be consistent with the reason causing $Prop_\exists(class_3)$ [remind (\ref{EC})].
\\[1ex]

Here is an example from Xynthia data.
Consider the causal link

$Expected(VLPress) \cause Expected(SWind)$

together with the following ontological links

$\left\{\begin{array}{rcl}
Hurri &\isaobj & SWind\vspace{1ex}\\
Hurri &\isaobj & NatDis.
\end{array}\right\}$\\

The links below can be obtained in the enriched model\\

$\left\{\begin{array}{rcl}
Expected(Hurri) & \isaont & Expected(SWind)\\
Expected(Hurri) & \isaont & Expected(NatDis)
\end{array}\right\}$\\[3ex]

\begin{minipage}{.45\textwidth}We get an argument to the effect that from the set of data
 {\boldmath$\Theta$} =\\
\noindent$\left\{\hspace{-1.3ex}\begin{array}{cl}Expected(VLPress)\cause Expected(SWind)\\
Expected(Hurri) \isaont Expected(StrW)\\
Expected(Hurri)   \isaont   Expected(NatDis)\hspace{-1ex}\end{array}\hspace{-1.1ex}\right\}$\\

we obtain:\hfill $\Theta$  yields that \hfilld\\

\noindent$Expected(VLPress)$ {\em can explain}\\ $Expected(NatDis)$
provided  $Expected(Hurri)$ is possible.
\end{minipage}
\\[1ex]

The intuition is that, from these data, it is reasonable to explain $Expected(NatDis)$,
by $Expected(VLPress)$, \prov{$Expected(Hurri)$}  (not contradicted by other data).

\subsection{An example of compound explanations}\label{ssecextrexpl}
\begin{figure}[h!]\
\centerline{\includegraphics{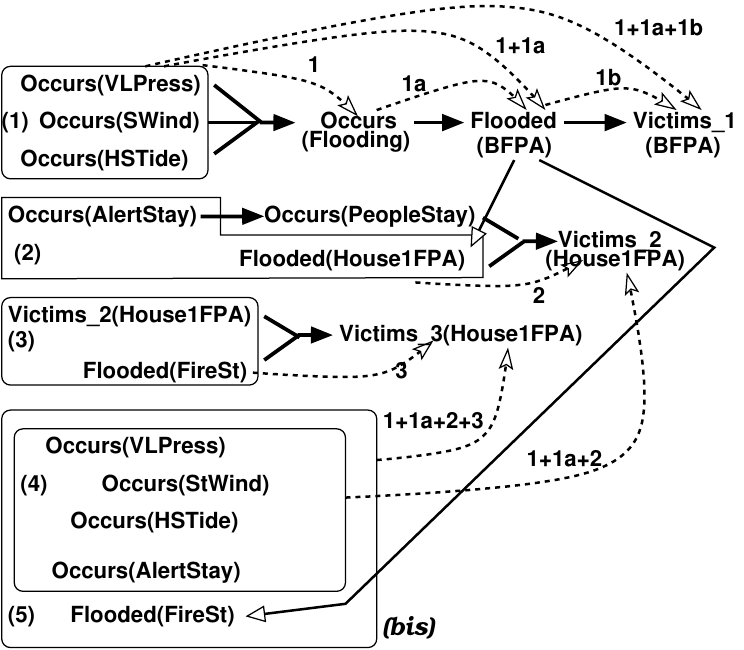}
}
\vspace*{-2ex}
\caption{A few explanations for victims\vspace*{-5ex}}\vspace*{-3ex}
\label{figvictims1}
\end{figure}

Figure \ref{figvictims1} displays an example from Xynthia (cf also Fig. \ref{figvictims0}) 
of a few possible explanations, represented by dotted lines, with their label such as $1$ or $1a$. The sets of literals, from which the explaining links start, are framed and numbered from (1) to (5). This shows transitivity of explanations at work: e.g.
 {\tt set}$\,1$ {\em can explain} {\tt Victims\_1(BFPA)} (explanation link labelled $1\!+\!1a\!+\!1b$) uses explanation links $1$, $1a$ and $1b$.\\[1ex]

Another example is \\
\indent {\tt set}$\,5$ {\em can explain}
 {\tt Victims\_3(House1FPA)} (explanation link $1$+$1a$+$2$+$3$):\\
explanations $1$, $1a$, $2$ and $3$ are at work here,
and link $2$ uses links $1$ and $1a$ together with\\
 {\tt Flooded(BFPA)} \isaont {\tt Flooded(House1FPA)}\\ while explanation $3$ comes from links $1$+$1a$ together with \\{\tt Flooded(BFPA)} \isaont {\tt Flooded(FireSt)}. 

\section{Argumentation}\label{secarg}

As just seen, the enriched causal graph allows us to infer explanations 
for assertions and these explanations might be used in an argumentative 
context \cite{BH08,Dung95}.
Let us first provide some motivation from our example.

A possible set of explanations for the flooded buildings is constituted by  the bad weather conditions (``very low pressure'' and ``strong wind'') together with ``high spring tide'' (see Fig.~\ref{gddessinC}). Given this explanation (argument), it is possible to 
attack it by noticing: a strong wind is supposed to trigger the red 
alarm of my anemometer and I did not get any alarm. However, this 
counter-argument may itself be attacked by remarking that, in the 
case of a hurricane, that is a kind of strong wind, an anemometer 
is no longer operating, which can explain that a red alarm cannot be observed. 

Let us see how to consider formally argumentation when relying on an 
enriched causal model and explanations as described in sections \ref{secenrich} and \ref{secexpl}.
Of course, we begin with introducing arguments, as follows.

\paragraph{Argument:}
That

``$\Phi$ can explain $\gamma$ in view of {\boldmath $\Theta$}, provided $\delta$ is possible''

\noindent is formalized here as an argument whenever (\ref{eqC1}) holds, that is:

\noindent {\boldmath $\Theta$}~yields that $\gamma$ can be explained by $\Phi$, provided $\delta$ is possible.\\

The components of an argument consist of:
\begin{itemize}
\item
$\Phi$, \emph{the explanation}, a set of literals.
\item
$\gamma$, \emph{the statement being explained}, a literal.
\item
$\Delta$, \emph{the justification} of the explanation (see Section \ref{secexpl}), a set of literals.
\item
{\boldmath $\Theta$}, \emph{the evidence}, \unboldmath
comprised of propositions (e.g., {$\big(\bigwedge\Phi\big)\rightarrow\gamma$)}, causal links\\ (e.g., $\Phi\cause\beta$), 
and ontological deduction links (e.g., $\delta \isaont \beta$).\\ 
\end{itemize}

Here is an illustration from the Xynthia event.
From Fig.~\ref{gddessinC}, that the $BFPA$ buildings are flooded can be explained via the set of two causal links\\

\noindent{\boldmath$\Theta = \{$ } \unboldmath
$Occurs(Flooding)  \cause  Flooded(BFPA),$ \\
\makebox[1em]{}$\left\{\hspace{-1ex}\begin{array}{c}Occurs(VLPress)\\Occurs(SWind)\\Occurs(HSTide)\end{array}\hspace{-1ex}\right\}\hspace{-.5ex} \cause  Occurs(Flooding)$
{\boldmath$\}$}\unboldmath\\

More precisely, using the basic case (\ref{eqbasic}) twice, we obtain that $Flooded(BFPA)$ can be explained by the set of literals 
\[
\Phi = \left\{\begin{array}{c}Occurs(VLPress)\\Occurs(SWind)\\Occurs(HSTide)\end{array}\right\}
\]
The corresponding argument is

{\boldmath $\Theta$}\unboldmath ~yields that $\gamma$ can be explained by $\Phi$, provided $\Delta$ is possible\\

where
\begin{itemize}
\item
\emph{The explanation} is $\Phi$. 
\item
\emph{The statement being explained} is $\gamma = Flooded(BFPA)$.
\item
\emph{The justification} of the explanation is empty.
\item
\emph{The evidence} is {\boldmath $\Theta$}.
\end{itemize}

\subsection{Counter-arguments}\label{sec:counterarguments}

Generally speaking, an argument ``$\Phi$ can explain $\gamma$ in view of {\boldmath $\Theta$}, provided $\Delta$ is possible'' is challenged by any statement which questions 
\begin{enumerate}
\item\label{x}
either $\Phi$ (e.g., an argument exhibiting an explanation for the negation of $\bigwedge\Phi$) 
\item\label{xx}
or $\gamma$ (e.g., an argument exhibiting an explanation for the negation of $\gamma$) 
\item\label{xxx}
or $\Delta$ (e.g., an argument exhibiting an explanation for the negation of $\bigwedge\Delta$) 
\item\label{xxxx}
or any item in {\boldmath $\Theta$} (e.g., an argument exhibiting an explanation for the negation of {\boldmath $\Theta$} for some $\theta$ occurring in {\boldmath $\Theta$})
\item\label{xxxxx}
or does so by refutation: using any of $\Phi$, {\boldmath $\Theta$}, 
$\Delta$ and $\gamma$ to explain some falsehood.
\end{enumerate}
Such objections are counter-arguments (they have the form of an argument: they explain something --but what they explain contradicts something in the challenged argument).

\paragraph{Dispute.}
Let us consider the illustration at the start of this section: 
The argument (that the buildings in the flood-prone area are flooded can be explained, partly, by a strong wind) is under attack on the grounds that my anemometer did not turn red --indicating that no strong wind occurred.
The latter is a counter-argument of type \ref{xxxxx} in above list. Indeed, 
the statement to be explained by the counter-argument is the falsehood $Red(anemo)$ (e.g., $Green(Anemo)$ has been observed), using $SWind$, i.e., an item in the explanation in the attacked argument.
Remember, the attacked argument is\\

{\boldmath$\Theta$} ~yields that $Flooded(BFPA)$ can be explained by \\
$\left\{\begin{array}{c}Occurs(VLPress)\\Occurs(SWind)\\Occurs(HSTide)\end{array}\right\}$\\

Taking $Red(Anemo)$ to be a falsehood, the counter-argument at hand is\\

{\boldmath $\Theta$}' ~yields that $Red(anemo)$ can be explained by \\
$\left\{\begin{array}{c}Occurs(SWind)\\OK(Anemo)\end{array}\right\}$\\

where
\begin{itemize}
\item
\emph{The explanation} is \\$\Phi' = \{Occurs(SWind), OK(Anemo)\}$.
\item
\emph{The statement being explained} is $\gamma' = Red(Anemo)$.
\item
\emph{The evidence} is \hspace{1em}
{\boldmath $\Theta$}$' = \\[1ex]
\left\{\left\{\begin{array}{c}Occurs(SWind)\\OK(Anemo)\end{array}\right\} \cause Red(anemo)\right\}$\\[1ex]
\end{itemize}

In our illustration, this counter-argument has in turn a counter-argument (of type \ref{x}.) explaining why the anemometer did not get red: 
i.e., explaining the negation of an item ($OK(Anemo)$ is the item in question) in the explanation in the counter-argument.
So, the counter-counter-argument here is: \hfill
{\boldmath $\Theta$}'' ~yields that\hfilld\\[1ex]
 $\neg OK(Anemo)$ can be explained by $\left\{\begin{array}{c}Occurs(Hurri)\end{array}\right\}$\\
where
\begin{itemize}
\item
\emph{The explanation} is $\Phi'' = \{Occurs(Hurri)\}$.
\item
\emph{The statement being explained} is \\ $\gamma'' = \neg OK(Anemo)$.
\item
\emph{The evidence} is
\end{itemize}

{\boldmath $\Theta$}$'' = \left\{ Occurs(Hurri) \cause \neg OK(Anemo) \right\}$\\

The dispute can extend to a counter-counter-counter-argument and so on as the process iterates.

\section{Conclusion}\label{secconcl}

The aim of this work is to study the link between causes and explanations \cite{HP01a,HP01b}, and to rely on explanations in an argumentative context \cite{BH08}.

In a first part, we define explanations as resulting from both causal and ontological links. An enriched causal model is built from a causal model and an ontology, from which the explaining links are derived (cf e.g. Fig\ref{figexPhilip1} and  (\ref{eqbasic}).
Our work differs from other approaches in the literature in that it strictly separates causality, ontology and explanations, while considering that ontology is key in generating sensible explanations from causal statements. Note however that some authors have already introduced ontology to be used for problem solving tasks as planning \cite[Chapter~2]{Kautz91} and more recently for diagnosis and repair \cite{Tortaetal08}.
We then argue that these causal explanations are interesting building blocks to be used in an argumentative context.

Although explanation and argumentation have long been identified as distinct processes \cite{Wal09}, it is also recognized that the distinction is a matter of context, hence they both play a role \cite{GibBroNun12} when it comes to eliciting an answer to a ``why'' question.
This is exactly what is attempted in this paper, as we are providing ``possible'' explanations, that thus can be turned into arguments.
The argument format has some advantages inasmuch as its uniformity allows us to express objection in an iterated way: ``possible'' explanations are challenged by counter-arguments that happen to represent rival, or incompatible, ``possible'' explanations.
However, a lot remains to be done.
Among others, comparing competing explanations according to minimality, preferences, and generally a host of criteria.

We have designed a system in answer set programming that implements most of the explicative proposal introduced above.
We plan to include it in an argumentative framework and think it will a good basis for a really practical
system, able to manage with a as rich and tricky example as the full Xinthia example.

\bibliography{iaf13causesbcm}

\end{document}